%% file: paper.tex
\documentclass[english]{lni}
\usepackage{booktabs}
\usepackage{multirow}
\usepackage{array}
\usepackage{tikz}
\usetikzlibrary{arrows.meta}

\newcommand{\benchmark}{Billiger.de Products}
\newcommand{\de}{DE}
\newcommand{\en}{EN}

\fancypagestyle{plain}{%
  \fancyhf{}%
  \fancyfoot[C]{\thepage}%
}

\begin{document}

\title[\benchmark]{\benchmark: A Bilingual Entity Matching Benchmark}

\author[1]{Aaron Steiner}{aaron.steiner@uni-mannheim.de}{0009-0006-6946-7057}
\author[1]{Ksenia Elagin}{ksenia.elagin@gmail.com}{}
\author[1]{Ralph Peeters}{ralph.peeters@uni-mannheim.de}{0000-0003-3174-2616}
\author[2]{Johannes Knopp}{jkn@solute.de}{}
\author[1]{Christian Bizer}{christian.bizer@uni-mannheim.de}{0000-0003-2367-0237}
\affil[1]{University of Mannheim\\Data and Web Science Group\\B6, 26\\68159 Mannheim\\Germany}
\affil[2]{solute GmbH\\Zeppelinstra\ss e 15\\76185 Karlsruhe\\Germany}
\maketitle

\begin{abstract}
Existing product matching benchmarks primarily contain English-language product data and are often dominated by a single product category, such as electronics. This paper introduces \benchmark{}, a bilingual German and English entity matching benchmark covering thirteen consumer product categories, including difficult-to-handle categories such as clothing and furniture. The benchmark data originates from the German price comparison platform billiger.de. Following the design of WDC Products, the benchmark offers multiple variants that differ in the fraction of corner cases, the size of the development set, and the fraction of entities unseen during training. An aligned English translation of every offer keeps all pairs, splits, and labels fixed, while cross-language test sets combine German and English records within individual pairs. We validate the benchmark using six supervised matchers and zero-shot GPT-5.2 on both language versions and the cross-language test sets. The validation shows the difficulty of the benchmark. The comparison of the results on the English version of the benchmark to the results on the German version shows that most matchers score on average higher on the English version. The difference is largest for RoBERTa and HierGAT, while the zero-shot LLM runs are largely insensitive to the language. Comparing the F1 scores achieved by PLM-based matchers on the English version of Billiger.de Products with their performance on existing English-language benchmarks, such as WDC Products and Abt-Buy, shows that \benchmark{} is more difficult than these benchmarks.
\end{abstract}


\input{sections/01_introduction}
\input{sections/03_benchmark}
\input{sections/04_matching_systems}
\input{sections/05_results}
\input{sections/06_cross_language}
\input{sections/02_related_work}
\input{sections/07_conclusion}

\clearpage
\section*{AI-Generated Content Acknowledgment}
Claude Opus 5 and GPT-5.6 supported code implementation for benchmark processing and evaluation (Sections~\ref{sec:benchmark}--\ref{sec:crosslanguage}) and language editing throughout the paper. The authors reviewed the suggestions and take responsibility for the research design, analysis, and conclusions.

As described in Sections~\ref{sec:english} and~\ref{sec:domains}, GPT-5-mini produced the English translations and assigned categories where the rule-based mapping left cases unresolved. DeepSeek V4 Flash, Gemini 3.8 Flash, and GLM 5.3 Flash screened translations as part of the audit, followed by human review.

\printbibliography

\end{document}

%% file: sections/01_introduction.tex
\section{Introduction}
\label{sec:introduction}

Entity matching is the task of discovering records that refer to the same real-world entity in a single or across multiple data sources~\cite{christen2012data,binette2022almost}. In e-commerce, the task appears at scale on price comparison platforms such as idealo.de, geizhals.de, or billiger.de, which continuously aggregate millions of product offers from thousands of shops and need to group offers that refer to the same product. Offers for the same product differ in naming conventions, attribute availability, attribute formatting, and description style. A shoe listed by one vendor as ``Nike Air Force 1 '07 -- White -- EU 42.5'' may appear on another platform as ``Nike AF1 Low '07 Sneaker -- Triple White -- US 9''.

Widely used entity matching benchmarks such as Abt-Buy, Amazon-Google, DBLP-Scholar, and Walmart-Amazon~\cite{koepcke2010evaluation,mudgal2018deep,primpeli2020profiling} provide a single level of task difficulty and contain almost exclusively English-language records. The WDC Products benchmark~\cite{peeters2023wdc} addressed the difficulty limitation through independently controllable variation along three dimensions: the amount of corner cases, the fraction of entities unseen during training, and the development set size. However, WDC Products also consists of English data and is dominated by consumer electronics. The matching systems evaluated in recent studies build on language models pre-trained predominantly on English text~\cite{li2020ditto,peeters2022supcon,yao2022hiergat}. How well such systems perform on non-English product data across a wide range of product categories is largely unknown. Existing benchmarks also do not cover the cross-language setting, in which product offers in different languages need to be matched, e.g. on a price comparison platform that aggregates offers across countries.

This paper presents \benchmark{}, an entity matching benchmark in German, the language of one of the largest e-commerce markets in Europe~\cite{lone2021ecommerce,sun2025bevh}. It is constructed from product offers provided by solute GmbH, the operator of the price comparison platform billiger.de.
Following WDC Products, it provides three development set sizes (small, medium, large), three corner-case ratios (20\,\%, 50\,\%, 80\,\%), and three proportions of unseen entities (0\,\%, 50\,\%, 100\,\%), yielding 27 variants (Section~\ref{sec:construction}). Its thirteen product categories include clothing and furniture, which are harder to handle for every evaluated matcher than electronics.

The English version contains GPT-5-mini translations of the name and description attribute values of every offer, with all pairs, splits, and labels kept fixed. Comparing the two versions measures language effects together with the translation, and pairing German records with English translations of their counterparts enables cross-language evaluation. We validate both versions using six supervised matchers and zero-shot GPT-5.2 with two prompt designs, and additionally evaluate XLM-R, a multilingual pretrained language model, on the 80\,\% corner-case variants and in the cross-language setting. The contributions are:

\begin{enumerate}
  \item An entity matching benchmark for German-language consumer product data, containing 13,568 offers that describe 2,168 products from 13 categories, with fixed splits along the three dimensions: amount of corner cases, fraction of entities unseen during training, and the development set size.
  \item An English translation of the complete benchmark that keeps all pairs, splits, and labels fixed, allowing direct comparability between entity matching in English and German as well as the construction of cross-language test sets.
  \item Validation results for six supervised matchers and zero-shot GPT-5.2 on all 27 variants of both language versions and on the cross-language test sets. The results show the difficulty of the benchmark in both languages. Under German-only training, four of the five PLM-based matchers lose performance on cross-language pairs compared to same-language pairs, while the scores of GPT-5.2 change by at most 1.4 F1.
\end{enumerate}

Section~\ref{sec:benchmark} describes benchmark construction, Section~\ref{sec:setup} introduces the matching systems, and Section~\ref{sec:results} reports their performance across difficulty dimensions, languages, and categories. Section~\ref{sec:crosslanguage} examines cross-language matching, followed by related work in Section~\ref{sec:related} and the conclusion in Section~\ref{sec:conclusion}.

The benchmark is released with the permission of solute GmbH and is available for public download.\footnote{\url{https://github.com/wbsg-uni-mannheim/billiger-de-products}} The released records contain no identifying information about shops, sellers, or persons.

%% file: sections/03_benchmark.tex
\section{Benchmark Creation}
\label{sec:benchmark}

This section describes benchmark construction from billiger.de product data, following WDC Products~\cite{peeters2023wdc} with controlled corner-case ratios, unseen-entity ratios, and development set sizes.

\subsection{Source Data}
\label{sec:source}

The raw data was provided by solute GmbH, the operator of the German price comparison platform billiger.de. 
Each offer carries the textual attributes \textit{name}, \textit{description}, and \textit{brand}, the numerical attribute \textit{price} in Euro, and identifying attributes such as EAN/GTIN and manufacturer part numbers (MPN) with varying coverage. The product assignments supplied by solute determine which offers refer to the same product. The matchers receive only name, description, brand, and price. The EAN/GTIN and MPN attributes are excluded, whereas identifiers inside names or descriptions remain, as the MPN in Figure~\ref{fig:cornercases} shows. This reflects matching tasks in which no separate identifier attributes are available.

Two preprocessing steps were applied to the source data: duplicate offers, caused for example by price updates of the same offer, were removed, as were offers whose names contain fewer than five words, as these are too sparsely described for meaningful matching. The remaining German-language offers cover the whole product range of a general-purpose price comparison portal (see Section~\ref{sec:domains}).

\subsection{Creation Process}
\label{sec:construction}

The benchmark realizes the same three dimensions as WDC Products~\cite{peeters2023wdc}: (i) the amount of corner cases, (ii) the fraction of unseen entities in the test set, and (iii) the development set size, meaning the combined size of the training and the validation split. Varying these dimensions supports systematic comparisons of matching performance. Corner cases are positives and negatives close to the decision boundary of the two classes. Figure~\ref{fig:cornercases} shows an example for a positive and a negative corner-case pair. A \textit{positive corner case} is a pair of matching offers whose surface forms are dissimilar, for example two offers for the same sneaker that describe the color using different vocabulary (Figure~\ref{fig:cornercases}, top). A \textit{negative corner case} is a pair of non-matching offers whose textual representations are highly similar, for example offers for two variants of the same wardrobe that agree in brand, model name, and nearly the entire description and differ only in some or a single attribute (Figure~\ref{fig:cornercases}, bottom). The corner-case ratio of a benchmark variant controls which share of its products contributes such pairs, and thus how hard the variant is. An \textit{unseen} entity is an entity, or product in this case, that appears in the test set but not in the training set, i.e. a trained matcher cannot directly memorize what separates this product from other similar products and instead needs to generalize to new products, analogous to the real world where new products are introduced daily.

\begin{figure}[tbp]
\scriptsize
\fbox{\begin{minipage}{0.97\textwidth}
\textbf{Positive corner case (label: match)}\\[2pt]
\textit{Offer A:} Brand: Puma \enspace Name: Puma \textbf{Gravition} 380738 22 \textbf{Pa Night/White/Vorange/Vgray}, Size: 42.5\\
Description: Puma Gravition 380738 22 Pa Night/White/Vorange/Vgray, Size: 42.5 \enspace Price: 65.00 Euro\\[3pt]
\textit{Offer B:} Brand: PUMA \enspace Name: PUMA \textbf{Graviton} Sneaker Regular Sneaker\\
Description: Size: 42.5, Size system: EU sizes, Color: \textbf{Parisian Night White Vibrant Orange Vaporous Gray} Blue, Upper material: Textile, Synthetic, Closure: Lace-up \enspace Price: 50.99 Euro
\end{minipage}}\\[3pt]
\fbox{\begin{minipage}{0.97\textwidth}
\textbf{Negative corner case (label: non-match)}\\[2pt]
\textit{Offer A:} Brand: RAUCH \enspace Name: rauch wardrobe \flqq Rasa\frqq{} beige \textbf{168 cm} x 188 cm x 52 cm\\
Description: [\dots] Color: light structured oak color/white [\dots] Number of shelves: \textbf{2}, Number of clothes rails: 1, Number of large drawers: \textbf{4}, Number of doors: \textbf{4 pcs.}, Type of doors: mirrored hinged doors, Width: \textbf{168 cm}, Depth: 52 cm, Height: 188 cm [\dots] \enspace Price: 409.99 Euro\\[3pt]
\textit{Offer B:} Brand: RAUCH \enspace Name: rauch wardrobe \flqq Rasa\frqq{} beige \textbf{85 cm} x 188 cm x 52 cm\\
Description: [\dots] Color: light structured oak color [\dots] Number of shelves: \textbf{1}, Number of clothes rails: 1, Number of large drawers: \textbf{2}, Number of doors: \textbf{2 pcs.}, Type of doors: mirrored doors, Width: \textbf{85 cm}, Depth: 52 cm, Height: 188 cm [\dots] \enspace Price: 230.99 Euro
\end{minipage}}
\caption{Two corner-case pairs in the English version. Top: matching offers for the same sneaker with a differently spelled model name and an abbreviated versus a written-out colorway. Bottom: non-matching offers for two wardrobe variants that differ only in width and interior configuration. Bold marks the decisive evidence, descriptions are truncated.}
\label{fig:cornercases}
\end{figure}

Figure~\ref{fig:benchmark-construction} summarizes the four construction steps: (i) grouping similar products, (ii) selecting products, (iii) splitting offers, and (iv) generating pairs.

The first step groups similar products to restrict the search for corner cases. The second selects 500 products for each corner-case ratio, combining similar but distinct products with randomly selected products. Disjoint product pools supply unseen evaluation products.

The third step splits offers into training, validation, and test sets. The proportion of products absent from training controls the unseen dimension, while the number of retained training offers controls development set size. The fourth step generates matching and non-matching pairs within these splits. Combining the three dimensions yields 27 variants.

After these four steps, the names and descriptions of all selected offers are translated into English while preserving the pairs, splits, and labels. This produces an aligned English counterpart for every German benchmark variant, as described in Section~\ref{sec:english}.

\begin{figure}[tbp]
\centering
\resizebox{\textwidth}{!}{\input{figures/benchmark-construction.tikz}}
\caption{Construction pipeline of the benchmark, from clustering to the aligned German and English releases.}
\label{fig:benchmark-construction}
\end{figure}

\paragraph{Grouping Similar Products.}
The cleansed corpus is first organized into coarse clusters of similar products using DBSCAN~\cite{ester1996dbscan} on the product name attribute. This restricts the subsequent search for similar products to smaller candidate sets.\footnote{Clustering code and parameters: \url{https://github.com/wbsg-uni-mannheim/billiger-de-products/blob/main/docs/benchmark-construction.md}.} Products with 7 to 80 offers form the pool for the seen products, as enough offers are needed for training, validation, and testing. Products with 4 to 6 offers form the pool for the unseen products, which contribute offers only to the evaluation splits, so seen and unseen products differ in offer count as well as in training exposure.

\paragraph{Product Selection.}
Each corner-case group contains five similar but distinct products selected from one DBSCAN cluster. One randomly chosen product serves as the anchor of the group. Four distinct products similar to the anchor are then added one by one. For each selection, the remaining candidates are ranked using one of four similarity metrics on the product name attribute: cosine, generalized Jaccard, Dice, or fastText embedding similarity. The metric is drawn at random for each selection so that no single metric dominates the process. The result is a group of five similar but distinct products, for example five wardrobes of the same model line in different widths. Offers belonging to these five products are used to construct corner-case pairs.

The three corner-case ratios combine corner-case products with random filler products drawn from the same candidate clusters. The 80\,\% set combines 400 corner-case products with 100 filler products, the 50\,\% set combines 250 with 250, and the 20\,\% set combines 100 with 400, so every set contains 500 products. The smaller corner-case selections are subsets of the larger ones, resulting in each set sharing part of their corner-case population. All development set sizes of one corner-case set contain the same 500 products and differ only in the number of offers retained per product.

\paragraph{Splitting.}
This step controls the splitting of the selected product offers into training, validation and test sets. For seen products, the offers are split into training, validation, and test sets, retaining at most 15 offers per product. Each offer is assigned to exactly one split of a configuration, so no offer and no pair occurs in more than one split. For each selected product, two offers go to test, two to validation, and the rest to training. The development set size dimension retains two (small), three (medium), or up to eleven (large) offers per product in the training split, with the smaller sets being subsets of the larger ones.

The unseen dimension is realized by replacing 0\,\%, 50\,\%, or 100\,\% of the products in the test split with products from an unseen pool of 500 additional products, selected by the same procedure and the same corner-case ratio. A third pool, constructed identically, provides the unseen products for the validation sets which are created in the same way as the unseen test sets. Following WDC Products, we call the 0\,\%, 50\,\%, and 100\,\% unseen conditions \textit{Seen}, \textit{Half-Seen}, and \textit{Unseen}, respectively.

\paragraph{Pair Generation.}
The final step turns the selected offers into labelled record pairs. Two offers are labelled as a match if they share the same product assignment supplied by solute. Positive pairs are generated exhaustively from such offers in each training, validation and test split. Additionally, for each offer, high similarity negative pairs are selected among the remaining 499 products using SoftTF-IDF, cosine, generalized Jaccard, or Dice similarity on the product name, one pair in the small, two in the medium, and three in the large configuration, plus one additional random negative pair. After these steps, the benchmark comprises nine training sets (3 corner-case ratios $\times$ 3 sizes), nine validation configurations in three sizes, and nine test sets (3 corner-case ratios $\times$ 3 unseen proportions), yielding overall 27 benchmark variants.

\subsection{English Version of the Benchmark}
\label{sec:english}

The English version of the benchmark contains translations of the attributes \textit{name} and \textit{description} of every record from German to English. All identifiers, product assignments, splits, labels, and the attributes \textit{brand} and \textit{price} stay unchanged. The records were translated split by split with the GPT-5-mini model (snapshot \texttt{gpt-5-mini-2025-08-07}), instructed to return JSON mirroring the record structure. Outputs violating the format were repaired or requested again. As a result, every offer in the benchmark exists in an aligned German and English version and every German benchmark variant has an identical English counterpart. Furthermore, German and English records can be combined into cross-language pairs for the experiments in Section~\ref{sec:crosslanguage}. The translation also keeps locale conventions unchanged, so prices stay in Euro and measurements in the metric system.

The translation was validated in two steps. Automated equality checks confirmed that identifiers, counts, and dataset dimensions equal the German version. For the content, three LLMs of different families, DeepSeek V4 Flash, Gemini 3.8 Flash, and GLM 5.3 Flash, compared the German and English versions of 3,000 randomly sampled test offers with the same prompt, checking identifiers, variant attributes, quantities, and omissions. An error counts as matching-relevant if it could change a match decision, for example a changed model number, size, material, or package quantity. One author then reviewed, blind to the model judgments, all 75 offers that at least one model flagged as matching-relevant, 100 offers that all models judged error-free, and 50 further offers, and found 11, zero, and one matching-relevant errors, respectively. Weighting these rates by stratum size yields an estimated 0.68\,\% matching-relevant translation errors. The matchers in Section~\ref{sec:language} achieve similar performance when trained on either language version further supporting that the translation has not negatively impacted the matching signals.

\subsection{Benchmark Statistics}
\label{sec:statistics}

Both language versions of the benchmark contain overall 13,568 product offers describing 2,168 products.\footnote{This paper describes benchmark version 1.1, released as tag \texttt{v1.1.0}: \url{https://github.com/wbsg-uni-mannheim/billiger-de-products/tree/v1.1.0}.} Table~\ref{tab:pairstats} reports the pair composition of the training, validation, and test splits. The validation columns show the Seen validation sets used for model selection, whose products all occur in the corresponding training sets. The test sets exhibit a class imbalance of roughly 1:10 to 1:13. To assess label quality, one author re-annotated a stratified random sample of 100 unique test pairs, comprising 50 matches and 50 non-matches, blind to the released labels. The re-annotation agreed with all 50 match labels and 49 of the 50 non-match labels. Weighting these class-specific disagreement rates by the 1:12.8 class imbalance of the unique test pairs yields an estimated disagreement rate of 1.9\,\%.

\begin{table}[htbp]
\centering\footnotesize
\setlength{\tabcolsep}{4pt}
\begin{tabular}{ll rrr rrr l rrr}
\toprule
 & & \multicolumn{3}{c}{\textbf{Training}} & \multicolumn{3}{c}{\textbf{Validation}} & \multicolumn{4}{c}{\textbf{Test}} \\
\cmidrule(lr){3-5}\cmidrule(lr){6-8}\cmidrule(lr){9-12}
\textbf{CC} & \textbf{Size} & \textbf{Pos.} & \textbf{Neg.} & \textbf{Total} & \textbf{Pos.} & \textbf{Neg.} & \textbf{Total} & \textbf{Unseen} & \textbf{Pos.} & \textbf{Neg.} & \textbf{Total} \\
\midrule
\multirow{3}{*}{20\,\%} & Small & 485 & 1,979 & 2,464 & 491 & 1,993 & 2,484 & 0\,\% & 407 & 4,038 & 4,445 \\
 & Medium & 1,457 & 4,456 & 5,913 & 491 & 2,989 & 3,480 & 50\,\% & 347 & 4,082 & 4,429 \\
 & Large & 13,100 & 14,012 & 27,112 & 491 & 3,984 & 4,475 & 100\,\% & 360 & 4,089 & 4,449 \\
\midrule
\multirow{3}{*}{50\,\%} & Small & 482 & 1,979 & 2,461 & 479 & 1,986 & 2,465 & 0\,\% & 363 & 4,049 & 4,412 \\
 & Medium & 1,464 & 4,441 & 5,905 & 479 & 2,977 & 3,456 & 50\,\% & 326 & 4,092 & 4,418 \\
 & Large & 13,024 & 14,000 & 27,024 & 479 & 3,971 & 4,450 & 100\,\% & 369 & 4,089 & 4,458 \\
\midrule
\multirow{3}{*}{80\,\%} & Small & 492 & 1,983 & 2,475 & 485 & 1,985 & 2,470 & 0\,\% & 336 & 4,036 & 4,372 \\
 & Medium & 1,462 & 4,435 & 5,897 & 485 & 2,970 & 3,455 & 50\,\% & 339 & 4,088 & 4,427 \\
 & Large & 12,747 & 13,824 & 26,571 & 485 & 3,967 & 4,452 & 100\,\% & 348 & 4,099 & 4,447 \\
\bottomrule
\end{tabular}
\caption{Pair composition per corner-case ratio (CC), identical for both languages. Validation reports the Seen splits used for model selection. Training and validation vary by development set size, while test sets vary by unseen-product ratio.}
\label{tab:pairstats}
\end{table}

The benchmark attributes do not contain many missing values as \textit{name} and \textit{price} are available in every product offer, whereas \textit{description} has a density of over 99\,\% and \textit{brand} of over 94\,\% in all variants. %

\subsection{Product Categories}
\label{sec:domains}

The benchmark spans the full product catalog of a general-purpose price comparison portal. To enable category-level analysis, all products were assigned to a unified taxonomy of thirteen top-level categories.\footnote{Category vocabulary with German and English labels: \url{https://github.com/wbsg-uni-mannheim/billiger-de-products/blob/main/docs/product-taxonomy.json}.} The taxonomy provides a common top-level vocabulary across retailer-specific category hierarchies. The retailer-provided category paths (over 190,000 distinct strings) follow no common standard. We map these paths to the fixed categories in two stages: rule-based keyword mapping over normalized category paths, followed by GPT-5-mini classification of the remaining cases into the same taxonomy. A manual validation on a random sample of 500 products found 89.0\,\% of the assignments correct, 7.4\,\% ambiguous (products plausibly belonging to two categories, such as protein powder as sports or food), and 3.6\,\% wrong, with the errors spread across categories.

\begin{table}[htbp]
\centering\small
\setlength{\tabcolsep}{2.8pt}
\begin{tabular}{l rr@{\hskip 6pt} l rr}
\toprule
\textbf{Category} & \textbf{Train\,(\%)} & \textbf{Test\,(\%)} & \textbf{Category} & \textbf{Train\,(\%)} & \textbf{Test\,(\%)} \\
\midrule
Clothing \& Accessories  & 20.85 & 28.56 & Cosmetics \& Drugstore & 3.25 & 4.51 \\
Furniture \& Living      & 21.10 & 26.42 & Health \& Care        & 2.24 & 2.46 \\
Electronics \& Computers & 20.53 & 12.51 & Office Supplies       & 0.94 & 0.69 \\
Tools \& DIY             & 11.34 & 8.10  & Food \& Beverages     & 1.00 & 0.68 \\
Sports \& Leisure        & 6.23  & 6.22  & Books, Films \& Music & 0.49 & 0.51 \\
Toys \& Baby             & 6.24  & 4.92  & Pet Supplies          & 0.23 & 0.19 \\
Automotive \& Motorcycle & 5.56  & 4.32  &                       &      &      \\
\bottomrule
\end{tabular}
\caption{Share of product offers per category in the training and test splits in percent, averaged over all variants.}
\label{tab:domaindist}
\end{table}

Table~\ref{tab:domaindist} shows the resulting category distribution. Clothing \& Accessories, Furniture \& Living, and Electronics \& Computers account for the largest shares, which mirrors European e-commerce markets, where fashion is the most frequently purchased category, followed by home-related goods and electronics~\cite{lone2021ecommerce,sun2025bevh}.

%% file: figures/benchmark-construction.tikz
\begin{tikzpicture}[
  x=1cm,y=1cm,
  font=\sffamily\fontsize{8.5}{10.3}\selectfont,
  line width=.55pt, line cap=round, line join=round,
  panel/.style={draw=black!85,rounded corners=3pt},
  title/.style={font=\sffamily\bfseries\fontsize{9.8}{11.4}\selectfont,
    align=center,anchor=north,inner sep=0pt},
  bclabel/.style={align=center,inner sep=1pt},
  product/.style={draw,rectangle,fill=white,minimum size=2.2mm,inner sep=0pt},
  offer/.style={draw,circle,fill=white,minimum size=2mm,inner sep=0pt},
  flow/.style={-{Stealth[length=1.5mm,width=1.3mm]}},
  bcstep/.style={-{Stealth[length=1.8mm,width=2mm]},line width=1.1pt},
  splitbox/.style={draw,fill=white,minimum width=1.9cm,minimum height=.65cm,
    align=center,inner sep=2pt}
]
\ifcsname\string\color@benchmarkBlue\endcsname\else
  \definecolor{benchmarkBlue}{RGB}{66,133,244}
\fi
\foreach \left/\right in {0/2,2.2/4.45,4.65/7.65,7.85/11.65,11.85/14.55,14.75/17}
  \draw[panel] (\left,0) rectangle (\right,8.2);
\foreach \left/\right in {2/2.2,4.45/4.65,7.65/7.85,11.65/11.85,14.55/14.75}
  \draw[bcstep] (\left,4.1) -- (\right,4.1);

\node[title] at (1,7.94) {Product\\Catalog};
\node[title] at (3.325,7.94) {(i) Grouping\\Similar\\Products};
\node[title] at (6.15,7.94) {(ii) Product\\Selection};
\node[title] at (9.75,7.94) {(iii) Splitting};
\node[title] at (13.2,7.94) {(iv) Pair\\Generation};
\node[title] at (15.875,7.94) {Translation};

\draw (0.28,3.15) arc[start angle=180,end angle=360,x radius=.72,y radius=.24];
\draw (0.28,3.15) -- (0.28,5.7);
\draw (1.72,3.15) -- (1.72,5.7);
\foreach \y in {4,4.85}
  \draw (0.28,\y) arc[start angle=180,end angle=360,x radius=.72,y radius=.24];
\draw[fill=white] (1,5.7) ellipse[x radius=.72,y radius=.24];
\foreach \y in {3.45,4.3,5.15}
  \foreach \x in {.52,.91,1.3} \node[product] at (\x,\y) {};
\node[bclabel] at (1,1.65) {offers with\\product\\assignments};

\foreach \cx/\cy/\radius in {2.88/6.2/.49,3.78/5.05/.46,2.9/3.95/.45,3.72/2.8/.49}
{
  \draw (\cx,\cy) circle[radius=\radius];
  \foreach \dx/\dy in {-.18/.16,.18/.16,0/-.19}
    \node[product] at ({\cx+\dx},{\cy+\dy}) {};
}
\node[bclabel] at (3.325,.75) {DBSCAN on\\product names};

\draw[benchmarkBlue,line width=.9pt] (5.58,5.9) ellipse[x radius=.76,y radius=1.0];
\foreach \x/\y in {5.13/6.27,6.03/6.27,5.13/5.46,6.03/5.46}
{
  \draw[benchmarkBlue,line width=.8pt] (5.58,5.9) -- (\x,\y);
  \node[product] at (\x,\y) {};
}
\node[product,draw=benchmarkBlue,line width=1pt,minimum size=3mm] at (5.58,5.9) {};
\node[bclabel,text=black] at (5.58,6.62) {anchor};
\foreach \x/\y in {6.73/6.46,7.25/6.28,6.86/5.95,7.33/5.64,6.72/5.39,7.16/5.08}
  \node[product] at (\x,\y) {};
\node[bclabel] at (5.58,4.53) {corner-case\\group (5)};
\node[bclabel] at (7.01,4.53) {random\\filler};
\foreach \y/\ratio/\share in {3.65/{80/20}/.8,2.97/{50/50}/.5,2.29/{20/80}/.2}
{
  \node[bclabel,anchor=east] at (5.57,\y) {\ratio};
  \fill[black!15] (5.67,{\y-.17}) rectangle (7.4,{\y+.17});
  \fill[benchmarkBlue] (5.67,{\y-.17}) rectangle ({5.67+1.73*\share},{\y+.17});
  \draw (5.67,{\y-.17}) rectangle (7.4,{\y+.17});
  \draw ({5.67+1.73*\share},{\y-.17}) -- ({5.67+1.73*\share},{\y+.17});
}
\node[bclabel] at (6.15,1.72) {3 ratios,\\500 products each};
\node[bclabel,draw=black!55,dashed,rounded corners=2pt,
  minimum width=2.64cm,minimum height=1.1cm] at (6.15,.7)
  {separate unseen\\pools for test\\and validation};

\node[product] (splitproduct) at (8.2,6.67) {};
\draw[flow] (8.38,6.67) -- (8.72,6.67);
\foreach \x in {8.95,9.26,9.57,9.88} \node[offer] at (\x,6.67) {};
\node[bclabel] at (10.23,6.67) {$\cdots$};
\node[bclabel] at (9.75,7.17) {per product: at most 15 offers};
\node[splitbox] (train) at (10.38,5.71) {Train};
\node[splitbox] (validation) at (10.38,4.62) {Validation};
\node[splitbox] (test) at (10.38,3.53) {Test};
\draw (10.53,6.67) -- (11.43,6.67) -- (11.43,6.16) -- (8.52,6.16) -- (8.52,3.53);
\draw[flow] (8.52,5.71) -- node[bclabel,above] {rest} (train.west);
\draw[flow] (8.52,4.62) -- node[bclabel,above] {2} (validation.west);
\draw[flow] (8.52,3.53) -- node[bclabel,above] {2} (test.west);
\foreach \x/\size in {9.02/small,10.05/medium,11.1/large}
  \node[bclabel,draw=black!55,fill=black!5,rounded corners=1pt,
    inner xsep=2pt,inner ysep=1.5pt] at (\x,5.2) {\size};
\foreach \x in {8.23,8.55,8.87} \node[product] at (\x,2.82) {};
\node[bclabel] at (8.55,2.3) {unseen\\products};
\draw[flow] (9.06,2.82) -- (10.03,2.82) -- (10.03,3.205);
\draw (test.south) -- (10.38,1.86) -- (8.52,1.86);
\draw (10.38,1.86) -- (11.04,1.86);
\foreach \x/\name/\share in {8.52/Seen/0,9.78/Half-Seen/50,11.04/Unseen/100}
{
  \node[draw,minimum width=1.13cm,minimum height=.69cm,inner sep=1pt,
    align=center] (regime-\share) at (\x,1.28) {\name\\\share\,\%};
  \draw[flow] (\x,1.86) -- (regime-\share.north);
}
\node[bclabel] at (9.75,.5) {unseen products in test};

\draw[rounded corners=3pt] (12.38,6.27) rectangle (14.02,6.91);
\node[offer] (positive-a) at (12.72,6.59) {};
\node[offer] (positive-b) at (13.68,6.59) {};
\draw (positive-a) -- (positive-b);
\node[bclabel] at (13.2,5.79) {match\\within a product};
\draw[benchmarkBlue,line width=.8pt] (13.2,4.89) ellipse[x radius=1.06,y radius=.49];
\foreach \x in {12.64,13.76}
  \draw[rounded corners=1pt] ({\x-.23},4.64) rectangle ({\x+.23},5.14);
\node[offer] (hard-a) at (12.64,4.89) {};
\node[offer] (hard-b) at (13.76,4.89) {};
\draw (hard-a) -- (hard-b);
\node[bclabel] at (13.2,4.01) {non-match,\\similarity-based};
\foreach \x in {12.64,13.76}
  \draw[rounded corners=1pt] ({\x-.23},2.97) rectangle ({\x+.23},3.47);
\node[offer] (random-a) at (12.64,3.22) {};
\node[offer] (random-b) at (13.76,3.22) {};
\draw[dashed] (random-a) -- (random-b);
\node[bclabel] at (13.2,2.62) {random non-match};
\node[bclabel] at (13.2,1.04) {27 variants\\(3 corner-case ratios\\$\times$ 3 sizes\\$\times$ 3 test sets)};

\node[bclabel] at (15.875,6.85) {German};
\node[bclabel] at (15.875,3.69) {English\\translation};
\foreach \base in {5.22,2.06}
{
  \draw[fill=white] (15.2,{\base+.16}) rectangle (16.73,{\base+1.27});
  \draw[fill=white] (15.04,\base) rectangle (16.57,{\base+1.11});
  \foreach \column in {0,1,2,3}
    \foreach \row in {0,1,2}
      \draw[fill=black!5] ({15.16+.35*\column},{\base+.13+.32*\row})
        rectangle ++(.24,.22);
}
\draw[{Stealth[length=1.5mm]}-{Stealth[length=1.5mm]}]
  (15.51,4.03) -- (15.51,4.98);
\node[bclabel,anchor=west] at (15.62,4.51) {aligned};
\node[bclabel] at (15.875,.82) {GPT-5-mini,\\pairs, splits,\\and labels fixed};
\end{tikzpicture}

%% file: sections/04_matching_systems.tex
\section{Validation of the Benchmark}
\label{sec:setup}

We evaluate six supervised matchers and zero-shot GPT-5.2 on all 27 variants of both languages to assess benchmark difficulty and provide baselines. XLM-R is evaluated as an additional supervised matcher on the 80\,\% corner-case variants (Section~\ref{sec:language}) and in the cross-language experiment (Section~\ref{sec:crosslanguage}). We distinguish feature-based matchers, fine-tuned pretrained language model (PLM)-based matchers, and the zero-shot LLM. The PLM-based group comprises RoBERTa, XLM-R, R-SupCon, HierGAT, and Ditto.

Precision, recall, and F1 are measured for the match class. Supervised results are mean test F1 over three training seeds per variant. The standard deviation across seeds has a median of 1.1 F1 and stays below 3 F1 for 85\,\% of the supervised configurations. Larger deviations occur mainly for Ditto and HierGAT on the small development sets. Per-seed results, standard deviations, precision, and recall are available in the repository. Hyperparameters follow the WDC Products codebase\footnote{\url{https://github.com/wbsg-uni-mannheim/wdcproducts}}. The following paragraphs describe each matcher.

\textbf{WordCooc:} A symbolic baseline used in WDC Products~\cite{peeters2023wdc} that uses binary word co-occurrence statistics over brand, name, description, and price as features for classical classifiers (Logistic Regression, LinearSVC, Decision Trees, Random Forests, Naive Bayes, XGBoost), with hyperparameters from a randomized search. For each training set, the classifier with the highest validation F1 is reported.

\textbf{Magellan:} Magellan~\cite{konda2016magellan} automatically generates token-, character-, and numeric-similarity features for each attribute based on attribute statistics. Magellan trains the same classifiers as the WordCooc baseline on these features, with hyperparameter optimization on the validation set.

\textbf{RoBERTa:} A RoBERTa-base cross-encoder~\cite{liu2019roberta} that serializes each product offer pair into a combined input sequence and is fine-tuned end-to-end with a binary classification head and early stopping. RoBERTa is predominantly pre-trained on English data.

\textbf{XLM-R:} XLM-R~\cite{conneau2020xlmr} is a RoBERTa variant pre-trained with masked language modeling on monolingual text in 100 languages, including German. It is fine-tuned using the same pipeline as the RoBERTa matcher.

\textbf{R-SupCon:} R-SupCon~\cite{peeters2022supcon} performs supervised contrastive pre-training of a RoBERTa-base model using the respective benchmark development set so that offers of the same product form clusters in the embedding space. The model is then fine-tuned end-to-end together with a classification head on pairwise offers for the final matching decision.

\textbf{HierGAT:} HierGAT~\cite{yao2022hiergat} combines transformer text encoding with a hierarchical graph representation of tokens, attributes, and records. It uses contextualized token embeddings to encode individual attributes and compares corresponding attributes across a record pair. Graph attention then weights these attribute comparisons to form a representation of the pair, which a classifier uses to predict whether the records match.

\textbf{Ditto:} Ditto~\cite{li2020ditto} is a transformer cross-encoder based on RoBERTa-base and one of the first PLM-based systems for entity matching. Compared to the RoBERTa baseline, Ditto adds data augmentation, for which we use span deletion.

\textbf{GPT-5.2 (zero-shot):} Following the protocol of Peeters et al.~\cite{peeters2025llm},
GPT-5.2 (snapshot \texttt{gpt-5.2-2025-12-11}) is queried zero-shot (without using the training sets) for every test pair using the OpenAI Batch API in July and September 2026. %
Two different prompts are compared, which are shown in Figure~\ref{fig:prompts}. The \textit{Simple} prompt asks whether two product descriptions refer to the same real-world product and requests a Yes/No answer. The \textit{Rule-Guided} prompt additionally encodes the identity definition of the benchmark: variants differing in size, color, configuration, capacity, or package quantity are not matches, conflicts in identifying attributes must yield ``No'', missing attributes alone are not a conflict, and a match requires positive evidence of equivalence. Both prompts serialize brand, name, description, and price and were adapted to the benchmark language. Requests use API defaults for reasoning effort and sampling parameters. Answers are normalized to binary decisions using the requested language or 1/0. We restrict the evaluation to a single LLM, as covering all combinations of test set, language, and prompt already requires roughly 160,000 requests per model.

\begin{figure}[tbp]
\scriptsize
\fbox{\begin{minipage}{0.97\textwidth}
\textbf{Simple Prompt}\\[2pt]
Do these two product descriptions refer to the same real-world product?
Answer with Yes or No only.\\
Product $i$: Brand: \textless brand\textgreater{} Name: \textless name\textgreater{} Description: \textless description\textgreater{} Price: \textless price\textgreater{} Euro
\end{minipage}}\\[3pt]
\fbox{\begin{minipage}{0.97\textwidth}
\textbf{Rule-Guided Prompt}\\[2pt]
You are an expert product matcher. Your task is to decide whether two product records refer to the EXACT same product (same GTIN/SKU). Analyze the provided records carefully and return your decision as strict Yes or No.
CRITICAL: Product variants are NOT matches. Different sizes, colors, configurations, or package quantities are DIFFERENT products with different GTINs.
Guidelines: Yes ONLY if both records refer to the exact same product that would have the same GTIN or barcode. No if they are variants of the same product line such as different size, color, capacity, or configuration. No if core identifying attributes conflict, including model numbers, dimensions, capacity, color, or configuration. Missing attributes alone are NOT a conflict. A match requires positive evidence of equivalence. Respond ONLY with Yes or No.\\
Product $i$: Brand: \textless brand\textgreater{} Name: \textless name\textgreater{} Description: \textless description\textgreater{} Price: \textless price\textgreater{} Euro
\end{minipage}}
\caption{The two zero-shot prompts for GPT-5.2 in their English form, with the serialized records appended for $i \in \{1,2\}$. The German versions are on the benchmark website.}
\label{fig:prompts}
\end{figure}

%% file: sections/05_results.tex
\section{Experimental Results}
\label{sec:results}

Section~\ref{sec:baselineresults} reports performance along the three benchmark dimensions, Section~\ref{sec:language} compares languages, and Section~\ref{sec:domainresults} compares product categories. Section~\ref{sec:wdccomparison} relates the results to WDC Products and Abt-Buy.

\subsection{Results Along the Benchmark Dimensions}
\label{sec:baselineresults}

Table~\ref{tab:resultsde} reports F1 for the six supervised matchers and both GPT-5.2 prompts across the 27 German variants, and Table~\ref{tab:resultsdelta} the difference to the English version per variant.

The results underline the difficulty of the benchmark. The feature-based approaches WordCooc and Magellan average 41.67 and 42.04 F1 across the 27 German variants, which is too low to be of practical use (Table~\ref{tab:resultsde}). Neural matchers such as Ditto reach up to 86.66 F1 in seen configurations but remain strongly affected by unseen entities. Averaged across all corner-case ratios and development set sizes, Ditto loses 21.17 F1 from the Seen to the Unseen test sets (Table~\ref{tab:resultsde}).
Zero-shot GPT-5.2 is also not able to solve the benchmark, as its scores under the stronger Rule-Guided prompt remain between 77.99 and 84.88 F1 (Table~\ref{tab:resultsde}).

The matching systems show clear differences across the three dimensions of the benchmark. F1 decreases as the corner-case ratio grows from 20\,\% to 80\,\%, for Ditto for example from 86.66 to 74.52 F1 in the large+seen setting (Table~\ref{tab:resultsde}). F1 further decreases from the Seen to the Unseen test sets, with drops of 39.16, 25.75, and 26.88 F1 for R-SupCon, Ditto, and RoBERTa in the large+20\,\% corner-case setting, against 26.59, 9.60, and 9.16 F1 reported for WDC Products. The broader product coverage may contribute to these drops. F1 generally increases with the development set size, with individual medium-to-large results deviating from this trend, consistent with PLM-based matchers profiting from more training examples~\cite{peeters2023wdc}. Comparing the easiest configuration (large, 20\,\% corner cases, Seen) with the hardest (small, 80\,\% corner cases, Unseen), the losses range from 24.60 F1 for Magellan to 55.47 F1 for R-SupCon.

\begin{table}[tbp]
\centering\scriptsize
\setlength{\tabcolsep}{3.3pt}
\begin{tabular}{lll rrrrrr rr}
\toprule
 & & & \multicolumn{6}{c}{\textbf{Supervised}} & \multicolumn{2}{c}{\textbf{GPT-5.2 zero-shot}} \\
\cmidrule(lr){4-9}\cmidrule(lr){10-11}
\textbf{CC} & \textbf{Size} & \textbf{Test} & \textbf{WordCooc} & \textbf{Magellan} & \textbf{RoBERTa} & \textbf{R-SupCon} & \textbf{HierGAT} & \textbf{Ditto} & \textbf{Simple} & \textbf{Rule} \\
\midrule
\multirow{9}{*}{20\,\%} & \multirow{3}{*}{Small} & Seen & 52.89 & 54.30 & 74.54 & 60.93 & 61.67 & 76.23 & \textbf{86.27} & \underline{80.29} \\
 &  & Half & 42.40 & 47.35 & 59.14 & 44.11 & 54.80 & 61.43 & \underline{82.98} & \textbf{84.11} \\
 &  & Unseen & 31.66 & 45.84 & 51.54 & 38.67 & 47.91 & 54.75 & \underline{82.00} & \textbf{84.88} \\
\cmidrule(lr){2-11}
 & \multirow{3}{*}{Medium} & Seen & 68.32 & 56.34 & 82.15 & 78.04 & 78.78 & \underline{82.68} & \textbf{86.27} & 80.29 \\
 &  & Half & 53.57 & 51.64 & 64.33 & 52.85 & 66.21 & 65.12 & \underline{82.98} & \textbf{84.11} \\
 &  & Unseen & 35.52 & 48.82 & 54.87 & 42.81 & 57.03 & 57.03 & \underline{82.00} & \textbf{84.88} \\
\cmidrule(lr){2-11}
 & \multirow{3}{*}{Large} & Seen & 72.77 & 57.03 & 84.08 & 84.85 & 83.58 & \textbf{86.66} & \underline{86.27} & 80.29 \\
 &  & Half & 53.19 & 48.21 & 66.23 & 57.91 & 67.52 & 70.05 & \underline{82.98} & \textbf{84.11} \\
 &  & Unseen & 34.62 & 46.56 & 57.20 & 45.69 & 56.86 & 60.91 & \underline{82.00} & \textbf{84.88} \\
\midrule
\multirow{9}{*}{50\,\%} & \multirow{3}{*}{Small} & Seen & 44.16 & 40.79 & 58.29 & 56.41 & 35.59 & 52.90 & \underline{82.82} & \textbf{84.41} \\
 &  & Half & 33.78 & 38.89 & 46.58 & 38.73 & 34.37 & 43.21 & \underline{73.13} & \textbf{81.35} \\
 &  & Unseen & 27.01 & 38.36 & 42.77 & 32.33 & 32.42 & 40.75 & \underline{75.06} & \textbf{78.54} \\
\cmidrule(lr){2-11}
 & \multirow{3}{*}{Medium} & Seen & 56.59 & 45.56 & 72.60 & 67.56 & 61.48 & 75.27 & \underline{82.82} & \textbf{84.41} \\
 &  & Half & 43.89 & 43.49 & 54.65 & 44.54 & 51.73 & 59.79 & \underline{73.13} & \textbf{81.35} \\
 &  & Unseen & 29.16 & 42.21 & 48.26 & 34.73 & 46.77 & 53.53 & \underline{75.06} & \textbf{78.54} \\
\cmidrule(lr){2-11}
 & \multirow{3}{*}{Large} & Seen & 63.08 & 43.85 & 77.78 & 76.63 & 74.48 & 81.40 & \underline{82.82} & \textbf{84.41} \\
 &  & Half & 47.73 & 40.00 & 57.74 & 48.67 & 59.14 & 61.85 & \underline{73.13} & \textbf{81.35} \\
 &  & Unseen & 28.27 & 40.44 & 48.67 & 36.47 & 53.97 & 52.02 & \underline{75.06} & \textbf{78.54} \\
\midrule
\multirow{9}{*}{80\,\%} & \multirow{3}{*}{Small} & Seen & 35.87 & 33.41 & 47.91 & 52.65 & 34.11 & 49.71 & \underline{74.37} & \textbf{79.23} \\
 &  & Half & 30.60 & 32.49 & 39.55 & 37.33 & 29.82 & 42.26 & \underline{67.11} & \textbf{77.99} \\
 &  & Unseen & 23.80 & 32.43 & 37.31 & 29.38 & 33.11 & 39.83 & \underline{66.25} & \textbf{80.90} \\
\cmidrule(lr){2-11}
 & \multirow{3}{*}{Medium} & Seen & 45.27 & 35.99 & 64.39 & 63.09 & 58.46 & 67.52 & \underline{74.37} & \textbf{79.23} \\
 &  & Half & 36.02 & 34.24 & 49.30 & 42.07 & 48.55 & 54.15 & \underline{67.11} & \textbf{77.99} \\
 &  & Unseen & 24.74 & 34.29 & 43.73 & 31.75 & 49.53 & 48.08 & \underline{66.25} & \textbf{80.90} \\
\cmidrule(lr){2-11}
 & \multirow{3}{*}{Large} & Seen & 52.90 & 35.60 & 72.23 & 70.39 & 66.65 & \underline{74.52} & 74.37 & \textbf{79.23} \\
 &  & Half & 36.25 & 33.76 & 54.47 & 47.24 & 51.69 & 57.24 & \underline{67.11} & \textbf{77.99} \\
 &  & Unseen & 21.15 & 33.11 & 45.28 & 33.36 & 47.35 & 49.44 & \underline{66.25} & \textbf{80.90} \\
\bottomrule
\end{tabular}
\caption{F1 on the German version per corner-case ratio (CC), development set size, and test set. Bold: best, underline: second best per variant. Zero-shot GPT-5.2 is independent of the development set size.}
\label{tab:resultsde}
\end{table}

Ditto and HierGAT are the strongest supervised systems across most German variants, reaching 86.66 and 83.58 F1 on the easiest variant. The small-data variants at high corner-case ratios separate the PLM-based models most clearly: on the German Seen test sets with the small development set at 50\,\% and 80\,\% corner cases, Ditto, RoBERTa, and R-SupCon stay between 47.91 and 58.29 F1, while HierGAT falls to 35.59 and 34.11 F1 (Table~\ref{tab:resultsde}).

GPT-5.2 varies less across the benchmark dimensions compared to the supervised systems: its F1 spans 20.02 points across the German variants under the Simple prompt and 6.89 points under the Rule-Guided prompt, against 24.60 to 55.47 points for the supervised models (Table~\ref{tab:resultsde}). Part of this stability follows from the design, as the development set size and unseen dimension do not affect a zero-shot model, which relies only on its general purpose training, in the same way as the matchers trained on the benchmark. With the fixed Rule-Guided prompt, GPT-5.2 reaches the highest F1 in 24 of 27 German variants (Table~\ref{tab:resultsde}). The Simple prompt leads on the small and medium Seen variants at 20\,\% corner cases, and Ditto leads on the large Seen variant. A caveat applies to interpreting the results for the GPT-5 model: the underlying offers were publicly available online and may have appeared in its pre-training data. The pairs, splits, and labels, however, were created specifically for this benchmark and were not available online before its release.
Averaged over the nine German test sets, GPT-5.2 reaches 76.66 F1 with the Simple and 81.30 with the Rule-Guided prompt.

\subsection{Results on the English Version}
\label{sec:language}

\begin{table}[tbp]
\centering\scriptsize
\setlength{\tabcolsep}{3.3pt}
\begin{tabular}{lll rrrrrr rr}
\toprule
 & & & \multicolumn{6}{c}{\textbf{Supervised}} & \multicolumn{2}{c}{\textbf{GPT-5.2 zero-shot}} \\
\cmidrule(lr){4-9}\cmidrule(lr){10-11}
\textbf{CC} & \textbf{Size} & \textbf{Test} & \textbf{WordCooc} & \textbf{Magellan} & \textbf{RoBERTa} & \textbf{R-SupCon} & \textbf{HierGAT} & \textbf{Ditto} & \textbf{Simple} & \textbf{Rule} \\
\midrule
\multirow{9}{*}{20\,\%} & \multirow{3}{*}{Small} & Seen & $+4.23$ & $-2.55$ & $+2.56$ & $+2.44$ & $+12.41$ & $-5.71$ & $+0.20$ & $-2.21$ \\
 &  & Half & $+2.19$ & $-1.58$ & $+2.05$ & $-0.08$ & $+9.70$ & $-3.40$ & $-1.43$ & $-0.48$ \\
 &  & Unseen & $+5.72$ & $-1.73$ & $+4.43$ & $-0.11$ & $+11.10$ & $-2.31$ & $-0.84$ & $+0.15$ \\
\cmidrule(lr){2-11}
 & \multirow{3}{*}{Medium} & Seen & $-2.05$ & $-2.43$ & $+1.38$ & $-0.12$ & $-1.54$ & $+3.35$ & $+0.20$ & $-2.21$ \\
 &  & Half & $-1.47$ & $-3.34$ & $+0.91$ & $-0.92$ & $-1.23$ & $+4.53$ & $-1.43$ & $-0.48$ \\
 &  & Unseen & $+5.60$ & $-2.26$ & $+1.55$ & $-0.31$ & $+0.03$ & $+4.15$ & $-0.84$ & $+0.15$ \\
\cmidrule(lr){2-11}
 & \multirow{3}{*}{Large} & Seen & $+0.60$ & $-0.36$ & $+2.22$ & $+0.20$ & $+1.89$ & $+1.49$ & $+0.20$ & $-2.21$ \\
 &  & Half & $+1.48$ & $+1.19$ & $-0.89$ & $-0.68$ & $+1.91$ & $+0.73$ & $-1.43$ & $-0.48$ \\
 &  & Unseen & $+6.64$ & $+0.58$ & $-1.44$ & $+0.09$ & $+3.92$ & $+1.73$ & $-0.84$ & $+0.15$ \\
\midrule
\multirow{9}{*}{50\,\%} & \multirow{3}{*}{Small} & Seen & $+2.36$ & $-1.42$ & $+7.15$ & $+4.60$ & $+14.02$ & $+9.64$ & $+0.54$ & $-1.98$ \\
 &  & Half & $+2.92$ & $-1.14$ & $+7.27$ & $+2.35$ & $+11.38$ & $+7.94$ & $-0.07$ & $+1.43$ \\
 &  & Unseen & $+3.59$ & $-2.61$ & $+6.79$ & $+1.10$ & $+10.40$ & $+6.08$ & $-1.10$ & $+3.40$ \\
\cmidrule(lr){2-11}
 & \multirow{3}{*}{Medium} & Seen & $-1.36$ & $-0.90$ & $+4.98$ & $+3.36$ & $+11.15$ & $-10.25$ & $+0.54$ & $-1.98$ \\
 &  & Half & $+2.11$ & $-0.10$ & $+4.42$ & $-1.24$ & $+11.32$ & $-3.92$ & $-0.07$ & $+1.43$ \\
 &  & Unseen & $+4.48$ & $-1.85$ & $+5.54$ & $+0.23$ & $+12.07$ & $-2.86$ & $-1.10$ & $+3.40$ \\
\cmidrule(lr){2-11}
 & \multirow{3}{*}{Large} & Seen & $+0.34$ & $-1.32$ & $+2.43$ & $+2.40$ & $+3.61$ & $+0.91$ & $+0.54$ & $-1.98$ \\
 &  & Half & $+1.42$ & $-1.09$ & $+1.86$ & $+1.29$ & $+5.21$ & $+2.54$ & $-0.07$ & $+1.43$ \\
 &  & Unseen & $+6.45$ & $-2.65$ & $+2.48$ & $+0.51$ & $+4.45$ & $+3.39$ & $-1.10$ & $+3.40$ \\
\midrule
\multirow{9}{*}{80\,\%} & \multirow{3}{*}{Small} & Seen & $+4.42$ & $-1.26$ & $+8.59$ & $+1.49$ & $+15.03$ & $+10.87$ & $-0.26$ & $+0.36$ \\
 &  & Half & $+4.42$ & $-0.88$ & $+5.82$ & $+2.19$ & $+13.64$ & $+9.56$ & $+1.99$ & $+2.55$ \\
 &  & Unseen & $+0.57$ & $-0.99$ & $+4.77$ & $+1.70$ & $+10.64$ & $+8.84$ & $+1.27$ & $+0.53$ \\
\cmidrule(lr){2-11}
 & \multirow{3}{*}{Medium} & Seen & $+4.44$ & $-2.69$ & $+5.19$ & $+1.04$ & $+6.28$ & $-10.76$ & $-0.26$ & $+0.36$ \\
 &  & Half & $+3.98$ & $-2.27$ & $+8.35$ & $+0.69$ & $+5.77$ & $-4.50$ & $+1.99$ & $+2.55$ \\
 &  & Unseen & $+1.46$ & $-2.45$ & $+5.59$ & $-0.05$ & $+2.34$ & $-3.18$ & $+1.27$ & $+0.53$ \\
\cmidrule(lr){2-11}
 & \multirow{3}{*}{Large} & Seen & $-1.67$ & $-2.58$ & $+1.42$ & $-0.89$ & $+3.97$ & $-0.07$ & $-0.26$ & $+0.36$ \\
 &  & Half & $+1.28$ & $-1.32$ & $+5.36$ & $-3.33$ & $+7.81$ & $+4.38$ & $+1.99$ & $+2.55$ \\
 &  & Unseen & $+2.19$ & $+0.31$ & $+7.05$ & $-1.31$ & $+7.31$ & $+4.92$ & $+1.27$ & $+0.53$ \\
\midrule
\multicolumn{3}{l}{Mean, small}  & $+3.38$ & $-1.57$ & $+5.49$ & $+1.74$ & $+12.04$ & $+4.61$ & $+0.03$ & $+0.42$ \\
\multicolumn{3}{l}{Mean, medium}  & $+1.91$ & $-2.03$ & $+4.21$ & $+0.30$ & $+5.13$ & $-2.60$ & $+0.03$ & $+0.42$ \\
\multicolumn{3}{l}{Mean, large}  & $+2.08$ & $-0.80$ & $+2.27$ & $-0.19$ & $+4.45$ & $+2.22$ & $+0.03$ & $+0.42$ \\
\midrule
\multicolumn{3}{l}{Mean, all 27} & $\mathbf{+2.46}$ & $\mathbf{-1.47}$ & $\mathbf{+3.99}$ & $\mathbf{+0.62}$ & $\mathbf{+7.21}$ & $\mathbf{+1.41}$ & $\mathbf{+0.03}$ & $\mathbf{+0.42}$ \\
\multicolumn{3}{l}{Median, all 27} & $+2.19$ & $-1.42$ & $+4.43$ & $+0.23$ & $+7.31$ & $+1.73$ & $-0.07$ & $+0.36$ \\
\multicolumn{3}{l}{Variants \en{} ahead} & 23 & 3 & 25 & 16 & 25 & 17 & 12 & 18 \\
\bottomrule
\end{tabular}
\caption{F1 difference between the English and the German version (\en{}$-$\de{}) per variant, positive values favour English. Adding a cell to Table~\ref{tab:resultsde} gives the English F1. Summary rows: mean by development set size, overall mean (bold) and median, and count of variants favouring English.}
\label{tab:resultsdelta}
\end{table}

This section presents the effect of the offer language on matching performance. For this experiment, supervised models were trained and evaluated on the English variant of the benchmark and then compared to the German results. Table~\ref{tab:resultsdelta} reports the per-variant F1 differences between the English and the German version for all evaluated systems. WordCooc and Magellan remain below 46 mean F1 in both languages, which is too low for practical matching applications, and their small mean language differences do not alter that conclusion. Averaged across all variants, every PLM-based matcher scores higher on the English version, which is likely due to their predominantly English pre-training data. The LLM GPT-5.2 is nearly indifferent to the language (Table~\ref{tab:resultsdelta}). Language sensitivity is larger when little training data is available: across the six supervised matchers, the mean absolute language gap decreases from 5.24 F1 with the small development sets to 3.52 with the medium and 2.40 with the large sets.

\paragraph{Lexical and Feature-Based Methods.} WordCooc shows a slight advantage on English in the Seen test sets, although the differences are too small to support a clear conclusion, whereas on the Unseen test sets the advantage is more visible, reaching 6.64 F1 points (large, 20\,\% corner cases, Table~\ref{tab:resultsdelta}). A possible explanation is that German compounds reduce exact token overlap between matching records~\cite{ziering2016compound}. Magellan instead slightly favours German (mean $-1.47$ F1).

\paragraph{PLM-based Models.} RoBERTa favours English with a mean difference of $+3.99$ F1, consistent with its predominantly English pre-training corpus~\cite{liu2019roberta}. The advantage is largest with small development sets and shrinks as more training data becomes available. Fine-tuning therefore allows the English-pretrained model to adapt to German product data. HierGAT shows the largest mean English advantage ($+7.21$ F1), while Ditto and R-SupCon are close to parity, although all three fine-tune the same RoBERTa backbone (Table~\ref{tab:resultsdelta}). 

\paragraph{Multilingual Encoder.} Table~\ref{tab:xlmr} compares the multi-lingual XLM-R with RoBERTa on the 80\,\% corner-case variants. On German, XLM-R scores above RoBERTa in all nine cells, by 1.9 to 9.6 F1, with the largest gains on the small development sets. On English the two stay within 3.8 F1 of each other, with RoBERTa ahead on the large sets. The average language difference is small with the multilingual backbone: over the nine cells, XLM-R scores 0.69 F1 higher on English than on German, RoBERTa 5.79 F1. This indicates that the English advantage of RoBERTa stems mainly from its backbone rather than from differences between the German and the translated English data, since the same pipeline on the same pairs shows almost no such advantage with the multilingual encoder. For matching in a non-English market, a multilingual encoder is the better starting point: it gains 1.9 to 9.6 F1 on German at a cost of at most 3.2 F1 on the large English sets.

\begin{table}[tbp]
\centering\small
\setlength{\tabcolsep}{6pt}
\begin{tabular}{ll rr rr}
\toprule
 & & \multicolumn{2}{c}{\textbf{German}} & \multicolumn{2}{c}{\textbf{English}} \\
\cmidrule(lr){3-4}\cmidrule(lr){5-6}
\textbf{Size} & \textbf{Test} & \textbf{RoBERTa} & \textbf{XLM-R} & \textbf{RoBERTa} & \textbf{XLM-R} \\
\midrule
\multirow{3}{*}{Small} & Seen & 47.91 & 57.54 & 56.50 & 57.51 \\
 & Half & 39.55 & 48.07 & 45.37 & 49.13 \\
 & Unseen & 37.31 & 43.31 & 42.07 & 45.23 \\
\midrule
\multirow{3}{*}{Medium} & Seen & 64.39 & 68.66 & 69.59 & 71.77 \\
 & Half & 49.30 & 55.65 & 57.65 & 57.96 \\
 & Unseen & 43.73 & 50.26 & 49.32 & 52.80 \\
\midrule
\multirow{3}{*}{Large} & Seen & 72.23 & 74.09 & 73.65 & 72.75 \\
 & Half & 54.47 & 58.01 & 59.83 & 56.90 \\
 & Unseen & 45.28 & 51.36 & 52.33 & 49.09 \\
\bottomrule
\end{tabular}
\caption{F1 of RoBERTa and XLM-R on the 80\,\% corner-case variants of both language versions, mean over three seeds.}
\label{tab:xlmr}
\end{table}

\begin{table}[tbp]
\centering\scriptsize
\setlength{\tabcolsep}{2.1pt}
\begin{tabular}{l rr rrrrrr rr}
\toprule
 & \multicolumn{2}{c}{\textbf{Records}} & \multicolumn{6}{c}{\textbf{Supervised}} & \multicolumn{2}{c}{\textbf{GPT-5.2}} \\
\cmidrule(lr){2-3}\cmidrule(lr){4-9}\cmidrule(lr){10-11}
\textbf{Category} & \textbf{Train} & \textbf{Test} & \textbf{WordCooc} & \textbf{Magellan} & \textbf{RoBERTa} & \textbf{R-SupCon} & \textbf{HierGAT} & \textbf{Ditto} & \textbf{Simple} & \textbf{Rule} \\
\midrule
Furniture \& Living               & 11,946 & 2,760 & 36.88 & 38.27 & 49.46 & 44.85 & 48.35 & 50.66 & 71.25 & \textbf{77.08} \\
Electronics \& Computers          & 11,737 & 885 & 49.55 & 58.19 & 77.82 & 67.34 & 77.40 & 80.95 & \textbf{90.81} & 90.76 \\
Clothing \& Accessories           & 9,100 & 2,514 & 38.20 & 33.35 & 49.48 & 43.35 & 46.96 & 52.06 & 64.76 & \textbf{70.94} \\
Tools \& DIY                      & 6,206 & 643 & 50.90 & 55.81 & 70.39 & 62.83 & 71.46 & 72.27 & \textbf{90.25} & 88.05 \\
Toys \& Baby                      & 3,754 & 367 & 48.80 & 51.74 & 76.70 & 70.24 & 72.80 & 75.88 & \textbf{92.53} & 88.52 \\
Sports \& Leisure                 & 3,711 & 737 & 42.31 & 45.50 & 59.74 & 52.94 & 57.80 & 63.81 & 81.83 & \textbf{89.59} \\
Automotive \& Motorcycle          & 3,044 & 293 & 43.54 & 50.60 & 64.75 & 50.06 & 62.10 & 65.38 & 86.48 & \textbf{91.74} \\
Cosmetics \& Drugstore            & 1,178 & 282 & 39.01 & 49.14 & 61.22 & 50.49 & 59.43 & 64.69 & \textbf{77.82} & 77.21 \\
Health \& Care                    & 1,057 & 222 & 56.42 & 67.91 & 79.70 & 68.31 & 81.04 & 80.26 & \textbf{95.50} & 82.72 \\
Food \& Beverages                 & 648 & 72 & 55.66 & 44.28 & 69.95 & 69.78 & 67.95 & \textbf{70.36} & 65.37 & 60.61 \\
Office Supplies                   & 491 & 58 & 41.19 & 56.94 & 79.48 & 71.57 & 78.62 & 81.64 & \textbf{88.89} & 75.56 \\
Books, Films \& Music             & 233 & 21 & 62.00 & 70.96 & 87.56 & 90.96 & 77.38 & 85.01 & \textbf{96.00} & 60.00 \\
\bottomrule
\end{tabular}
\caption{Mean same-category F1 on German over nine test sets, seeds, and development sizes. Bold: best per category. Train/Test count record occurrences in pairs of the large 80\,\% corner-case training/Half-Seen test sets. Pet Supplies has no records in this test set and is omitted. Books, Films \& Music has fewer than 50 test records.}
\label{tab:domainf1}
\end{table}

\paragraph{Large Language Model.} The largest individual difference for GPT-5.2, 3.40 F1 under the Rule-Guided prompt, stems from single runs per variant, and Section~\ref{sec:crosslanguage} reports an observed run-to-run spread of 1.10 F1 for repeated DE-DE requests. A small language effect under the Rule-Guided prompt therefore remains plausible, possibly because explicit rule application profits from the stronger representation of English in LLM pre-training corpora~\cite{qin2025multilingual}.

\subsection{Results Across Product Categories}
\label{sec:domainresults}

This section compares matching difficulty across the thirteen product categories. For each model, every test pair is annotated with the categories of its two products and F1 is computed per category pair. Table~\ref{tab:domainf1} reports the same-category F1, averaged over all nine German test sets, together with the number of records per category in the pairs of the 80\,\% corner-case large training set and the corresponding Half-Seen test set. Rows are ordered by the number of training records.

The category ranking is stable across the model families. Among the nine largest categories, Clothing \& Accessories and Furniture \& Living are the two hardest for every model. Within these nine categories, Clothing is the single hardest for five of the eight model columns. One possible explanation is that similar product variants, generally less structured attributes, and vendor-specific naming of colors, materials, and styles make identity harder to distinguish from textual similarity. Electronics \& Computers ranks among the three easiest categories for every model and Health \& Care for seven of the eight model columns, while Tools \& DIY and Toys \& Baby are in the upper half for every model. Model numbers, manufacturer identifiers, and standardized technical specifications may provide more distinctive matching signals in these categories.

The category matching performance ranking differs from the ranking by training-record counts. The three largest categories contain the two hardest categories as well as one of the easiest, and Health \& Care, with 1,057 training records against more than 9,000 in each of the three largest categories, scores above Clothing and Furniture for every model and above Electronics for six of the eight model columns (Table~\ref{tab:domainf1}). The lower scores for clothing and furniture show the value of evaluating matchers beyond electronics. Repeating the category analysis on the English benchmark version for WordCooc, Magellan, RoBERTa, R-SupCon, and GPT-5.2 yields the same picture: the ranking of the categories stays largely unchanged, and the per-category shifts are of the same order as each matcher's overall language difference (Section~\ref{sec:language}). The corresponding English category scores are available on the benchmark website.\footnote{\url{https://wbsg-uni-mannheim.github.io/billiger-de-products/\#results}}

\subsection{Comparison with WDC Products and Abt-Buy}
\label{sec:wdccomparison}

This section compares the English results with the published results of the same matchers on WDC Products~\cite[Table~3]{peeters2023wdc} and Abt-Buy~\cite[Table~3]{peeters2022supcon}. Averaged over the 27 configurations that \benchmark{} shares with WDC Products, RoBERTa, Ditto, HierGAT, and R-SupCon score 9.8 to 19.6 F1 lower on the English version of \benchmark{} (Tables~\ref{tab:resultsde} and~\ref{tab:resultsdelta}). WordCooc scores the same and Magellan 5.1 F1 higher, so the gap is specific to the PLM-based matchers, while the ranking of the six matchers is the same on both benchmarks.

Abt-Buy has no difficulty variants. Even on the easiest English configuration (20\,\% corner cases, large, Seen), RoBERTa, Ditto, and R-SupCon score 1.2 to 8.7 F1 below their published Abt-Buy results. Averaged across all 27 variants of \benchmark{}, their scores are 28.0 to 43.1 F1 lower. The published results were obtained with the training protocols of their papers, so the comparison is descriptive. A possible explanation for the gap is the category mix, as electronics dominate WDC Products and Abt-Buy and are among the easier categories in Section~\ref{sec:domainresults}. These comparisons show lower F1 for the PLM-based matchers on the English version of \benchmark{} than on the two established benchmarks.

%% file: sections/06_cross_language.tex
\section{Cross-Language Matching}
\label{sec:crosslanguage}

This section evaluates German-trained matchers on pairs whose records are written in different languages, simulating a German price comparison portal expanding to English-language markets. Section~\ref{sec:language} evaluated the two languages separately.

\subsection{Experimental Setup}
\label{sec:crosssetup}

All supervised matchers are trained and selected using the training and validation splits, respectively, of the German 80\,\% corner-case large development set. Each selected model is then evaluated, without further training, on five variants of the German 80\,\% corner-case Half-Seen test set (4,427 pairs each). In \textbf{DE-DE} both records are the German originals, which is the default same-language matching task. In \textbf{DE-EN} and \textbf{EN-DE} the left or the right record is replaced by its English translation. \textbf{EN-EN} represents the case of generalizing a German-trained matcher to fully English pairs. \textbf{Mixed} combines the four language combinations in equal shares.

GPT-5.2 receives the German prompt in all five variants, and its zero-shot requests were repeated three times per prompt and variant. The supervised matchers report the mean over three seeds from training runs that are separate from the main evaluation but use the same training and validation splits. Differences between the DE-DE scores in Tables~\ref{tab:resultsde} and~\ref{tab:crosslanguage} reflect these separate runs. Within this experiment, language effects are measured relative to each model's own DE-DE baseline.

\subsection{Results}
\label{sec:crossresults}

\begin{table}[tbp]
\centering\small
\setlength{\tabcolsep}{6pt}
\begin{tabular}{l r rrrr}
\toprule
 & \textbf{DE-DE} & \multicolumn{4}{c}{\textbf{Difference to DE-DE}} \\
\cmidrule(lr){2-2}\cmidrule(lr){3-6}
\textbf{Model} & \textbf{F1} & \textbf{DE-EN} & \textbf{EN-DE} & \textbf{EN-EN} & \textbf{Mixed} \\
\midrule
WordCooc            & 36.2 & $-6.0$ & $-6.6$ & $-10.5$ & $-7.0$ \\
Magellan            & 33.8 & $-2.2$ & $-3.0$ & $-0.5$ & $-1.2$ \\
RoBERTa             & 53.3 & $-7.2$ & $-7.7$ & $-5.3$ & $-5.2$ \\
XLM-R               & 58.0 & $-4.8$ & $-5.6$ & $+0.3$ & $-1.4$ \\
R-SupCon            & 43.1 & $-0.8$ & $-1.8$ & $-5.8$ & $-2.2$ \\
HierGAT             & 51.4 & $-7.4$ & $-6.9$ & $-0.7$ & $-4.0$ \\
Ditto               & 55.3 & $-6.4$ & $-6.2$ & $-2.4$ & $-3.4$ \\
\midrule
GPT-5.2 Simple      & 68.4 & $+1.4$ & $-0.4$ & $+0.5$ & $+1.3$ \\
GPT-5.2 Rule-Guided & 79.6 & $-0.7$ & $-0.7$ & $-0.8$ & $-0.8$ \\
\bottomrule
\end{tabular}
\caption{F1 on the five language variants of the 80\,\% corner-case Half-Seen test set and difference to DE-DE, computed from unrounded means. Supervised matchers are trained on German pairs only and report means over seeds, GPT-5.2 means over three repetitions.}
\label{tab:crosslanguage}
\end{table}

RoBERTa, XLM-R, HierGAT, and Ditto lose 4.8 to 7.7 F1 on DE-EN and EN-DE, and each of them loses less on fully English pairs, where the differences range from a gain of 0.3 to a loss of 5.3 F1 (Table~\ref{tab:crosslanguage}). The separate-language evaluation (Section~\ref{sec:language}) shows that the models can match both German and English offers. Cross-language pairs may disrupt shared lexical cues learned from German pairs when only one record is translated. XLM-R loses 4.8 and 5.6 F1 on DE-EN and EN-DE but gains 0.3 F1 on EN-EN, so multilingual pre-training alone does not prevent the cross-language loss in this setting. R-SupCon is the exception: it stays within 1.8 F1 of DE-DE on cross-language pairs and loses most on EN-EN (5.8 F1). Its cross-language F1 remains below XLM-R and Ditto, so the smaller loss does not imply stronger absolute performance.

WordCooc loses most on EN-EN (10.5 F1), because its co-occurrence features require a token to appear in both records, and shared English tokens may lie outside the feature space fitted on German training pairs. Magellan stays within 3.0 F1 of DE-DE across variants, consistent with its similarity features being less affected by language (Table~\ref{tab:crosslanguage}).

GPT-5.2 reaches the highest F1 in every variant, and its scores differ from DE-DE by at most 1.4 F1. With both prompts, the spread across variants (0.77 F1 Rule-Guided, 1.85 F1 Simple) is smaller than the spread across three repetitions of identical DE-DE requests (1.10 and 1.98 F1). These comparisons describe the observed spread rather than establish statistical significance.

The EN-EN results support earlier observations of transfer between languages~\cite{peeters2022crosslanguage,alves2024crosslingual}. The cross-language test sets additionally expose a transfer difficulty for four of the five PLM-based matchers under German-only training.

%% file: sections/02_related_work.tex
\section{Related Work}
\label{sec:related}

\paragraph{Entity Matching Methods.} Entity matching has been researched since the 1940s~\cite{dunn1946record,fellegi1969theory}. Early systems compared records using string similarity metrics~\cite{cohen2003comparison}. Supervised systems such as Magellan generate similarity features across record attributes and train classifiers on them~\cite{konda2016magellan}. Christophides et al.~\cite{christophides2020overview} give an overview of the resolution workflows of this period. Deep learning architectures learn record representations end-to-end~\cite{mudgal2018deep} and are surveyed in~\cite{barlaug2021neural,karapiperis2026survey}. PLM-based systems include cross-encoders such as Ditto~\cite{li2020ditto}, supervised contrastive learning in R-SupCon~\cite{peeters2022supcon}, and hierarchical graph attention in HierGAT~\cite{yao2022hiergat}. Section~\ref{sec:setup} describes these systems. Recently, large language models have been applied to entity matching in zero-shot settings~\cite{narayan2022foundation,peeters2023chatgpt,peeters2025llm,zhang2025deepdive} as well as via fine-tuning~\cite{steiner2025finetuning}. Wang et al.~\cite{wang2025match} compare pairwise matching with comparing and selecting candidate records, finding selection the most cost-effective strategy in their experiments. These studies show that LLMs reach performance comparable to task-specific matchers without labelled training data. Almost all of them were conducted on English benchmarks.

\begin{table}[tbp]
\centering\scriptsize
\setlength{\tabcolsep}{2.5pt}
\begin{tabular}{l l l ccccc}
\toprule
\textbf{Benchmark} & \textbf{Domain} & \textbf{Size} & \textbf{Multiling.} & \textbf{Sizes} & \textbf{Corner c.} & \textbf{Unseen} & \textbf{Cross-lang.} \\
\midrule
Abt-Buy~\cite{koepcke2010evaluation}      & electronics, other goods & 9,575 pairs & -- & -- & -- & -- & -- \\
Walmart-Amazon~\cite{konda2016magellan}   & electronics & 10,242 pairs & -- & -- & -- & -- & -- \\
WDC Products~\cite{peeters2023wdc}        & electronics, other goods & 11,715 offers & -- & \checkmark & \checkmark & \checkmark & -- \\
Ember~\cite{wang2022ember}                & clothing and shoes & 126,277 records & -- & -- & -- & \checkmark & -- \\
Polish PM dataset~\cite{mozdzonek2022polish} & drugstore, beverages & 24,752 pairs & -- & \checkmark & -- & -- & -- \\
ProMapCz~\cite{mackova2023promap}         & consumer products & 1,495 pairs & -- & -- & -- & -- & -- \\
eCommerce~\cite{alves2024thesis}           & electronics & 69,740 pairs & -- & -- & -- & -- & -- \\
Notas Fiscais~\cite{alves2024thesis}       & dairy, cosmetics, parts & 37,832 pairs & -- & -- & -- & -- & -- \\
Markt-Pilot~\cite{rettenmeier2023marktpilot} & industrial products & $>$100,000 pairs & \checkmark & -- & -- & -- & -- \\
\midrule
\textbf{\benchmark{} (ours)}             & consumer products & 13,568 offers & \checkmark & \checkmark & \checkmark & \checkmark & \checkmark \\
\bottomrule
\end{tabular}
\caption{Product matching benchmarks. Multiling.: records in multiple languages. Sizes: multiple development set sizes. Corner c./Unseen: controlled corner-case/unseen-entity ratios. Cross-lang.: different languages within a pair.}
\label{tab:benchmarks}
\end{table}

\paragraph{Entity Matching Benchmarks.} A broad range of benchmarks supports the evaluation of matching systems. The tasks compiled by K{\"o}pcke et al.~\cite{koepcke2010evaluation}, such as Abt-Buy and Amazon-Google, and the Magellan datasets~\cite{konda2016magellan}, such as Walmart-Amazon, are widely used and were profiled in detail by Primpeli and Bizer~\cite{primpeli2020profiling}. The Alaska benchmark covers multi-source integration tasks for cameras and monitors~\cite{crescenzi2021alaska}, and Machamp compiles matching tasks over pairs of differing structuredness~\cite{wang2021machamp}. All of these benchmarks contain English records and provide a single level of task difficulty per dataset. Ember, a Chinese-language product matching benchmark, additionally evaluates generalization to unseen entities~\cite{wang2022ember}. Mo{\.z}d{\.z}onek et al.~\cite{mozdzonek2022polish} release the first open Polish product matching dataset, built from drugstore and beverage offers in three training set sizes, and ProMapCz provides Czech product pairs with graded close and medium non-matches~\cite{mackova2023promap}. Alves~\cite{alves2024thesis} describes two Portuguese corpora: eCommerce contains product titles from Brazilian marketplaces, and Notas Fiscais contains short invoice descriptions of dairy products, cosmetics, and vehicle suspension parts. Both provide random and hard-negative pair variants. Table~\ref{tab:benchmarks} reports their hard-negative variants, summing the three categories for Notas Fiscais~\cite[Table~6.3]{alves2024thesis}. The Markt-Pilot dataset provides a large collection of web-scraped offer pairs for industrial products such as automation and machine components in German and English~\cite{rettenmeier2023marktpilot}. WDC Products~\cite{peeters2023wdc} added the multi-dimensional design with controlled corner-case ratios, unseen-entity ratios, and three development set sizes, built from English schema.org data dominated by consumer electronics. Table~\ref{tab:benchmarks} compares the discussed benchmarks along language, domain, size, and difficulty dimensions. Our benchmark adopts the design of WDC Products and adds German-language consumer product data, a thirteen-category product mix, a parallel English translation, and cross-language test sets. The PLM-based matchers evaluated in Section~\ref{sec:results} score 10 to 20 F1 lower on the English version of \benchmark{} than on WDC Products. Even on the easiest English configuration, RoBERTa, Ditto, and R-SupCon score below their published Abt-Buy results, positioning \benchmark{} as a challenging benchmark for English-language product matching (Section~\ref{sec:wdccomparison}).

\paragraph{Cross-Language Matching.} A recent survey of deep learning for entity resolution names cross-language matching among the open research directions~\cite{karapiperis2026survey}. Peeters and Bizer~\cite{peeters2022crosslanguage} observed that a multilingual transformer trained on English pairs can transfer matching performance to other languages. Alves et al.~\cite{alves2024crosslingual} train product matchers on English data for use on Portuguese data and report promising results with little training data in the target language. In both studies, the language of a record is a property of its data source rather than a controlled condition. \benchmark{} provides the same pairs in German, English, and two cross-language versions. Combined with its controlled corner-case ratios, unseen-entity ratios, and development set sizes, this supports studying language effects alongside all three difficulty dimensions.

%% file: sections/07_conclusion.tex
\section{Conclusion}
\label{sec:conclusion}

This paper introduced \benchmark{}, an entity matching benchmark built from offers of the German price comparison platform billiger.de. It covers thirteen product categories, provides fixed splits along the dimensions corner-case ratio, fraction of unseen products, and development set size following the design of WDC Products, and includes an aligned English translation of every offer that keeps all pairs, splits, and labels fixed.

The evaluation of six supervised matchers and zero-shot GPT-5.2 on all 27 variants of both language versions shows that the benchmark is difficult: no supervised matcher exceeds 55 F1 in the 80\,\% corner-case Unseen variants. On the English version, the PLM-based matchers score 10 to 20 F1 below their published results on WDC Products. Even on the easiest English configuration, RoBERTa, Ditto, and R-SupCon score below their published Abt-Buy results, positioning \benchmark{} as a challenging benchmark for English-language product matching. Clothing and furniture are the hardest categories for every matcher and electronics among the easiest, which shows the value of evaluating matchers beyond electronics.

Averaged across variants, the PLM-based matchers score higher when trained and evaluated in English than in German, with the largest gains for RoBERTa and HierGAT. Under German-only training, RoBERTa, XLM-R, HierGAT, and Ditto lose 5 to 8 F1 on cross-language pairs, whereas R-SupCon stays close to its German score and GPT-5.2 changes by at most 1.4 F1. 
Future work can build on the benchmark in several directions. The cross-language test sets enable a systematic comparison of translate-then-match pipelines with multilingual matchers and LLMs.
Extending the evaluation to additional languages and models would test how far the stability observed for GPT-5.2 carries beyond German and English.